\documentclass[10pt,twocolumn,letterpaper]{article}

\usepackage{iccv}
\usepackage{times}
\usepackage{epsfig}
\usepackage{graphicx}
\usepackage{amsmath,amssymb}
\usepackage{commath}
\usepackage{textcomp}
\usepackage{enumitem} 

\usepackage{subcaption}
\usepackage{colortbl} 
\usepackage{xcolor}
\usepackage{soul}

\usepackage[pagebackref=true,breaklinks=true,letterpaper=true,colorlinks,bookmarks=false]{hyperref}

\iccvfinalcopy 

\def\iccvPaperID{531} 

\ificcvfinal\fi

\begin{document}
	
	\title{Binarized Convolutional Landmark Localizers for Human Pose Estimation and Face Alignment with Limited Resources}
	\author{Adrian Bulat  and 
		Georgios Tzimiropoulos\\
		Computer Vision Laboratory, The University of Nottingham\\
		Nottingham, United Kingdom\\
		{\tt\small \{adrian.bulat, yorgos.tzimiropoulos\}@nottingham.ac.uk}}
        
	\maketitle
	
	\begin{abstract}
 Our goal is to design architectures that retain the groundbreaking performance of CNNs for landmark localization and at the same time are lightweight, compact and suitable for applications with limited computational resources. To this end, we make the following contributions: (a) we are the first to study the effect of neural network binarization on localization tasks, namely human pose estimation and face alignment. We exhaustively evaluate various design choices, identify performance bottlenecks, and more importantly propose multiple orthogonal ways to boost performance. (b) Based on our analysis, we propose a novel hierarchical, parallel and multi-scale residual architecture that yields large performance improvement over the standard bottleneck block while having the same number of parameters, thus bridging the gap between the original network and its binarized counterpart. (c) We perform a large number of ablation studies that shed light on the properties and the performance of the proposed block. (d) We present results for experiments on the most challenging datasets for human pose estimation and face alignment, reporting in many cases state-of-the-art performance. Code can be downloaded from \url{https://www.adrianbulat.com/binary-cnn-landmarks}
	       
	\end{abstract}
	
	\section{Introduction} \label{sec:intro}
    
    	    	\begin{figure}[!htb]
		\begin{center}
			\begin{subfigure}[t]{0.19\textwidth}
				\centering
				\includegraphics[height=2.0in,trim={0.5cm 0.5cm 0.5cm 0.5cm},clip]{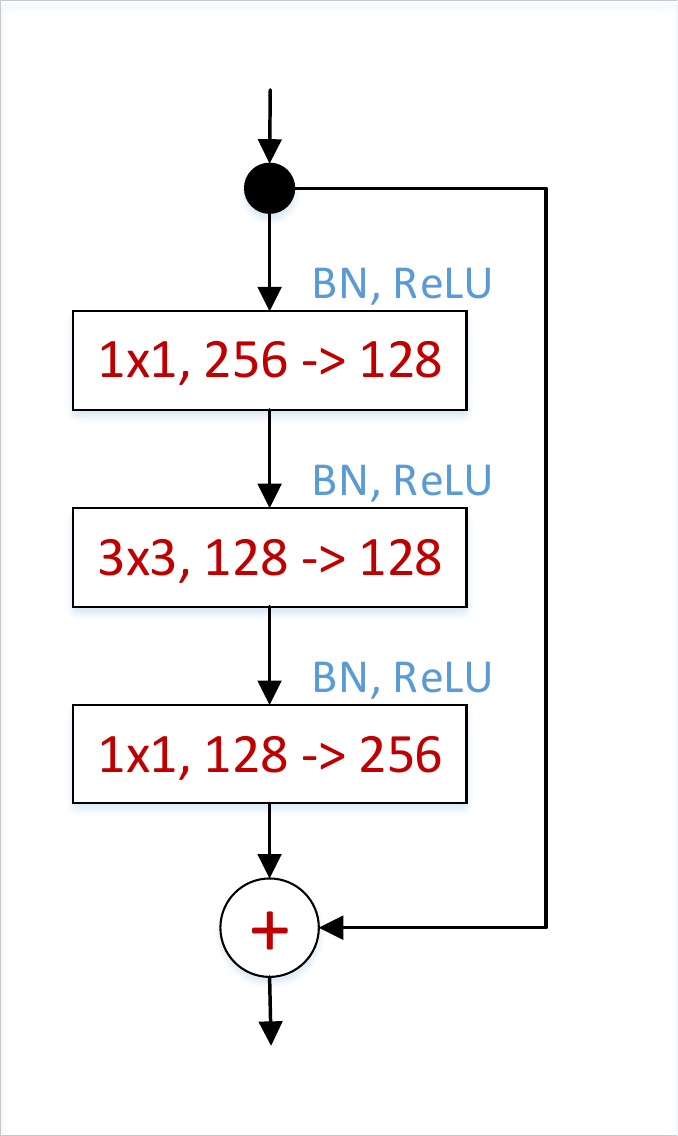}
				\caption{original}
				\label{fig:layers_original_intro}
			\end{subfigure}%
			~ 
			\begin{subfigure}[t]{0.29\textwidth}
				\centering
				\includegraphics[height=2.0in,trim={0.5cm 0.5cm 0.5cm 0.5cm},clip]{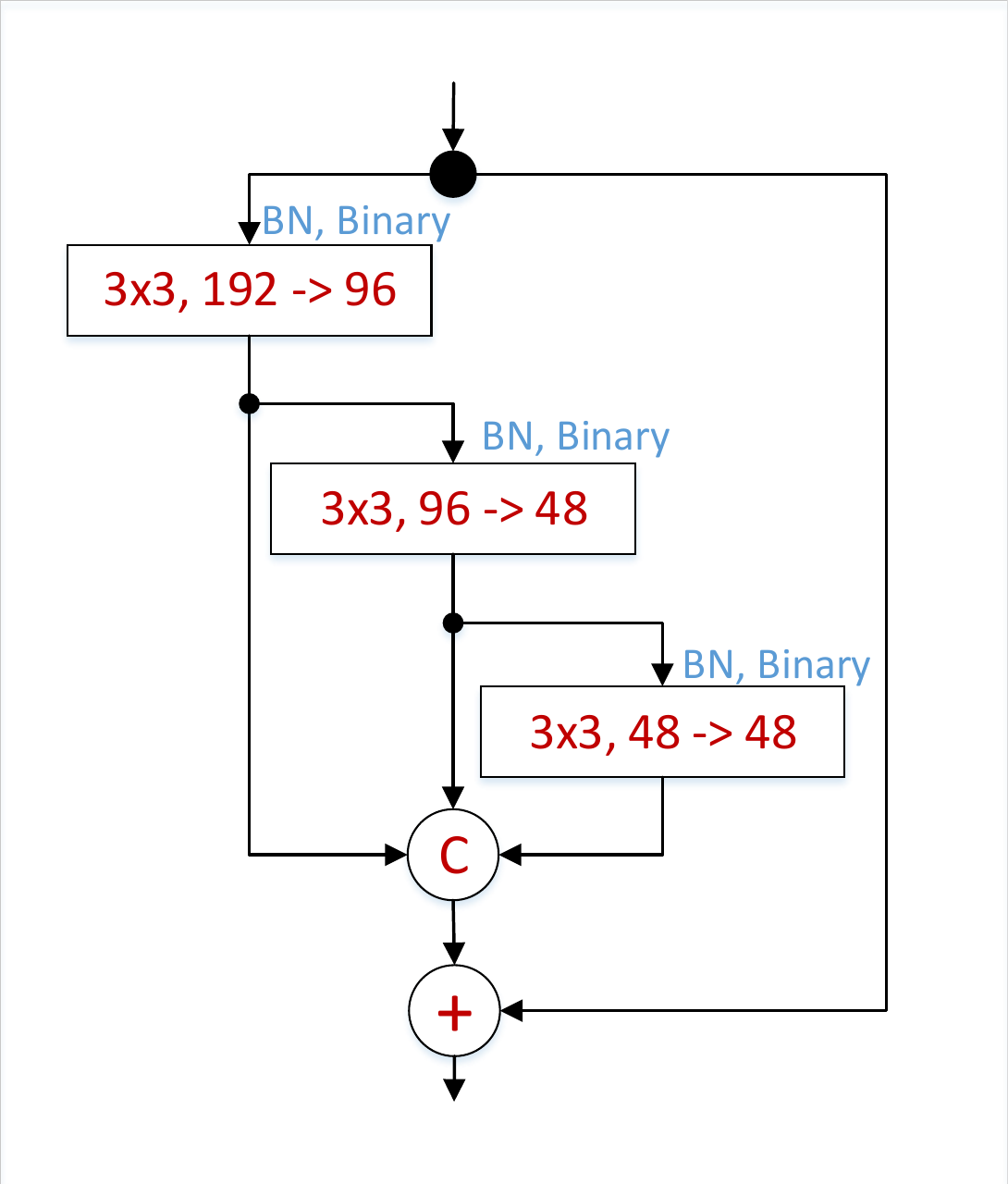}
				\caption{proposed}
				\label{fig:layers_wider_intro}
			\end{subfigure}
		\end{center}
		\caption{ (a) The original bottleneck layer of \cite{he2016identity}. (b) The proposed hierarchical parallel \& multi-scale structure: our block increases the receptive field size, improves gradient flow, is specifically designed to have (almost) the same number of parameters as the original bottleneck, does not contain $1 \times 1$ convolutions, and in general is derived from the perspective of improving the performance and efficiency for binary networks. \textbf{Note:} a layer is depicted as a rectangular block containing: its filter size, the number of input and output channels; "C" - denotes concatenation and "+" an element-wise sum.}
		\label{fig:intro_layers}
	\end{figure}
	This work is on localizing a predefined set of fiducial points on objects of interest which can typically undergo non-rigid deformations like the human body or face. Very recently, work based on Convolutional Neural Networks (CNNs) has revolutionized landmark localization, demonstrating results of remarkable accuracy even on the most challenging datasets for human pose estimation \cite{ bulat2016human,newell2016stacked,wei2016convolutional} and face alignment \cite{bulat2016two}. However, deploying (and training) such methods is computationally expensive, requiring one or more high-end GPUs, while the learned models typically require hundreds of MBs, thus rendering them completely unsuitable for real-time or mobile applications. This work is on highly accurate and robust yet efficient and lightweight landmark localization using binarized CNNs.  
    
	Our work is inspired by very recent results of binarized CNN architectures on image classification \cite{rastegari2016xnor, courbariaux2016binarized}. Contrary to these works, we are the first to study the effect of neural network binarization on fine-grained tasks like landmark localization. Similarly to \cite{rastegari2016xnor, courbariaux2016binarized}, we find that binarization results in performance drop, however to address this we opted to investigate and propose several architectural innovations which led to the introduction of a completely novel hierarchical, parallel and multi-scale residual block, as opposed to investigating ways to improve the binarization process as proposed in \cite{rastegari2016xnor, courbariaux2016binarized}.  In summary, \textbf{our contributions} are:

	\begin{enumerate}[leftmargin=*]
\setlength\itemsep{-0.35em}
		\item
		We are the first to study the effect of binarization on state-of-the-art CNN architectures for the problem of localization, namely human pose estimation and face alignment. To this end, we exhaustively evaluate various design choices, and identify performance bottlenecks. More importantly, we describe multiple orthogonal ways to boost performance; see Subsections \ref{ssec:wide_block}, \ref{ssec:size_filters} and \ref{ssec:conv1}. 
		\item
		Based on our analysis, we propose a new hierarchical, parallel and multi-scale residual architecture (see Subsection \ref{ssec:better}) specifically designed to work well for the binary case. Our block results in large performance improvement over the baseline binary residual block of \cite{he2016identity} (about 6\% in absolute terms when the same number of parameters are used (see Subsection \ref{sec:binary}, Tables~\ref{tab:binary_comp1} and~\ref{tab:binary_comp2})). 
		\item
		While our newly proposed block was developed with the goal of improving the performance of binary networks, we also show that the performance boost offered by the proposed architecture also generalizes to some extent for the case of real-valued networks (see Subsection \ref{sec:real}). 
		\item
		We perform a large number of ablation studies that shed light on the properties and the performance of the proposed block (see Sections \ref{sec:binaryvsreal} and \ref{sec:results}). 
        \item
        We present results for experiments on the most challenging datasets for human pose estimation and face alignment, reporting in many cases state-of-the-art performance (see Section \ref{sec:results}).
		
	\end{enumerate}

	
\section{Closely Related Work} \label{sec:related}
	
	This Section reviews related work on network quantization, network design, and gives an overview of the state-of-the-art on human pose estimation and face alignment. \newline 
	\textbf{Network quantization.} Prior work \cite{holi1993finite} suggests that high precision parameters are not essential for obtaining top results for image classification. In light of this, \cite{courbariaux2014training, lin2015fixed} propose 16- and 8-bit quantization, showing negligible performance drop on a few small datasets \cite{krizhevsky2009learning}. \cite{zhou2016dorefa} proposes a technique which allocates different numbers of bits (1-2-6) for the network parameters, activations and gradients. 
	
	Binarization (i.e. the extreme case of quantization) was long considered to be impractical due to the destructive property of such a representation \cite{courbariaux2014training}. Recently \cite{soudry2014expectation} showed this not to be the case and that by quantizing to $\{-1,1\}$ good results can be actually obtained. \cite{courbariaux2015binaryconnect} introduces a new technique for training CNNs that uses binary weights for both forward and backward passes, however, the real parameters are still required during training.  The work of \cite{courbariaux2016binarized} goes one step further and binarizes both parameters and activations. In this case multiplications can be replaced with elementary binary operations \cite{courbariaux2016binarized}. By estimating the binary weights with the help of a scaling factor, \cite{rastegari2016xnor} is the first work to report good results on a large dataset (ImageNet). Notably, our method makes use of the recent findings from \cite{rastegari2016xnor} and \cite{courbariaux2016binarized} using the same way of quantizing the weights and replacing multiplications with bit-wise \textit{xor} operations. 
	
	Our method differs from all aforementioned works in two key respects: (a) instead of focusing on image classification, we are the first to study neural network binarization in the context of a fine-grained computer vision task namely landmark localization (human pose estimation and facial alignment) by predicting a dense output (heatmaps) in a fully convolutional manner, and (b) instead of enhancing the results by improving the quantization method, we follow a completely different path, by enhancing the performance via proposing a novel architectural design for a hierarchical, parallel and multi-scale residual block. \newline

\textbf{Block design.} The proposed method uses a residual-based architecture and hence the starting point of our work is the \textit{bottleneck} block described in \cite{he2016deep,he2016identity}. More recently, \cite{xie2016aggregated} explores the idea of increasing the cardinality of the residual block by splitting it into a series of $c$ parallel (and much smaller so that the number of parameters remains roughly the same) sub-blocks with the same topology which behave as an ensemble. Beyond bottleneck layers, Szegedy \etal \cite{szegedy2015going} propose the inception block which introduces parallel paths with different receptive field sizes and various ways of lowering the number of parameters by factorizing convolutional layers with large filters into smaller ones. In a follow-up paper \cite{szegedy2017inception}, the authors introduce a number of inception-residual architectures. The latter work is the most related one to the proposed method.  
	
Our method is different from the aforementioned architectures in the following ways (see Fig.~\ref{fig:layers_wider_intro}): we create a hierarchical, parallel and multi-scale structure that (a) increases the receptive field size inside the block and (b) improves gradient flow, (c) is specifically designed to have (almost) the same number of parameters as the original bottleneck, (d) our block does not contain $1 \times 1$ convolutions, and (e) our block is derived from the perspective of improving the performance and efficiency of binary networks. \newline
\textbf{Network design.} Our target was not to propose a new network architecture for landmark localization; hence we used the state-of-the-art \textit{Hour-Glass} (HG) network of \cite{newell2016stacked} which makes use of the bottleneck block of \cite{he2016deep}. Because we are interested in efficiency, all of our experiments are conducted using a single network i.e. we do not use stacking as in \cite{newell2016stacked}. Our baseline was the binary HG obtained by directly quantizing it using \cite{rastegari2016xnor}. As Table~\ref{tab:bootlneck_vs} shows, there is a significant performance gap between the binary and the real valued HGs. We bridge this gap by replacing the bottleneck block used in the original HG with the proposed block. \newline
\textbf{Human Pose Estimation.} Recent work using CNNs has shown remarkable results \cite{toshev2014deeppose,tompson2014joint,pfister2015flowing,insafutdinov2016deepercut,bulat2016human,newell2016stacked,wei2016convolutional}, yet all these methods are computationally demanding, requiring at least one high-end GPU. In contrast, our network uses binary weights and activations and as such is intended to run on systems with limited resources (e.g. smartphones). \newline
\textbf{Face alignment.} Current state-of-the-art for large pose 2D and 3D face alignment is also based on CNNs \cite{jourabloo2016large,bulat2016convolutional,bulat2016two}; however, these methods are computationally demanding. Our network produces state-of-the-art results for this task, yet it is designed to run on devices with limited resources.
	
\section{Background} \label{sec:method}
	
	The ResNet consists of two type of blocks: \textit{basic} and \textit{bottleneck}. We are interested only in the latter one which was designed to reduce the number of parameters and keep the network memory footprint under control. We use the ``pre-activation'' version of \cite{he2016identity}, in which batch normalization \cite{ioffe2015batch} and the activation function precede the convolutional layer. This block is shown in Fig.~\ref{fig:layers_original_intro}. Note that we used the version of bottleneck defined in \cite{newell2016stacked} the middle layer of which has 128 channels (vs 64 used in \cite{he2016identity}). 
	
	The residual block is the main building block of the HG which is a state-of-the-art architecture for landmark localization that predicts a set of heatmaps (one for each landmark) in a fully convolutional fashion. The HG network is an extension of \cite{long2015fully} allowing however for a more symmetric top-down and bottom-up processing. See also \cite{newell2016stacked}.  
	
\section{Method} \label{S:Method}
	
Herein, we describe how we derive the proposed binary hierarchical, parallel and multi-scale block of Fig. \ref{fig:layers_parallel}. In Section~\ref{sec:binary}, by reducing the number of its parameters to match the ones of the original bottleneck, we further derive the block of Fig. \ref{fig:layers_wider_intro}. This Section is organized as follows: 
	\begin{itemize}[leftmargin=*]
    \setlength\itemsep{-0.35em}
		\item
		We start by analyzing the performance of the binarized HG in Subsection \ref{ssec:naive} which provides the motivation as well as the baseline for our method. 
		\item
		Then, we propose a series of architectural innovations in Subsections \ref{ssec:wide_block}, \ref{ssec:size_filters}, \ref{ssec:conv1} and \ref{ssec:better} (shown in Figs. \ref{fig:layers_wider}, \ref{fig:layers_mscale} and \ref{fig:layers_no1x1}) each of which is evaluated and compared against the binarized residual block of Subsection \ref{ssec:naive}. 
		\item
		Finally, by combining ideas from these architectures, we propose the binary hierarchical, parallel and multi-scale block of Fig. \ref{fig:layers_parallel}. Note that the proposed block is not a trivial combination of the aforementioned architectures but a completely new structure. 
	\end{itemize}
	
	We note that all results for this Section were generated for the task of human pose estimation   using the standard training-validation partition of MPII \cite{bulat2016human,newell2016stacked}. 
	
	\subsection{Binarized HG} \label{ssec:naive}
    
	We start from the original bottleneck blocks of the HG network and, following \cite{rastegari2016xnor}, we binarize them keeping only the first and last layers of the network real. This is crucial, especially for the very last layer where higher precision is required for producing a dense output (heatmaps). Note that these layers account for less than 0.01\% of the total number of parameters.

\begin{figure}[!htb]
		\begin{center}
			\centering
			\includegraphics[height=1.5in,trim={0.5cm 0.5cm 0.5cm 0.5cm},clip]{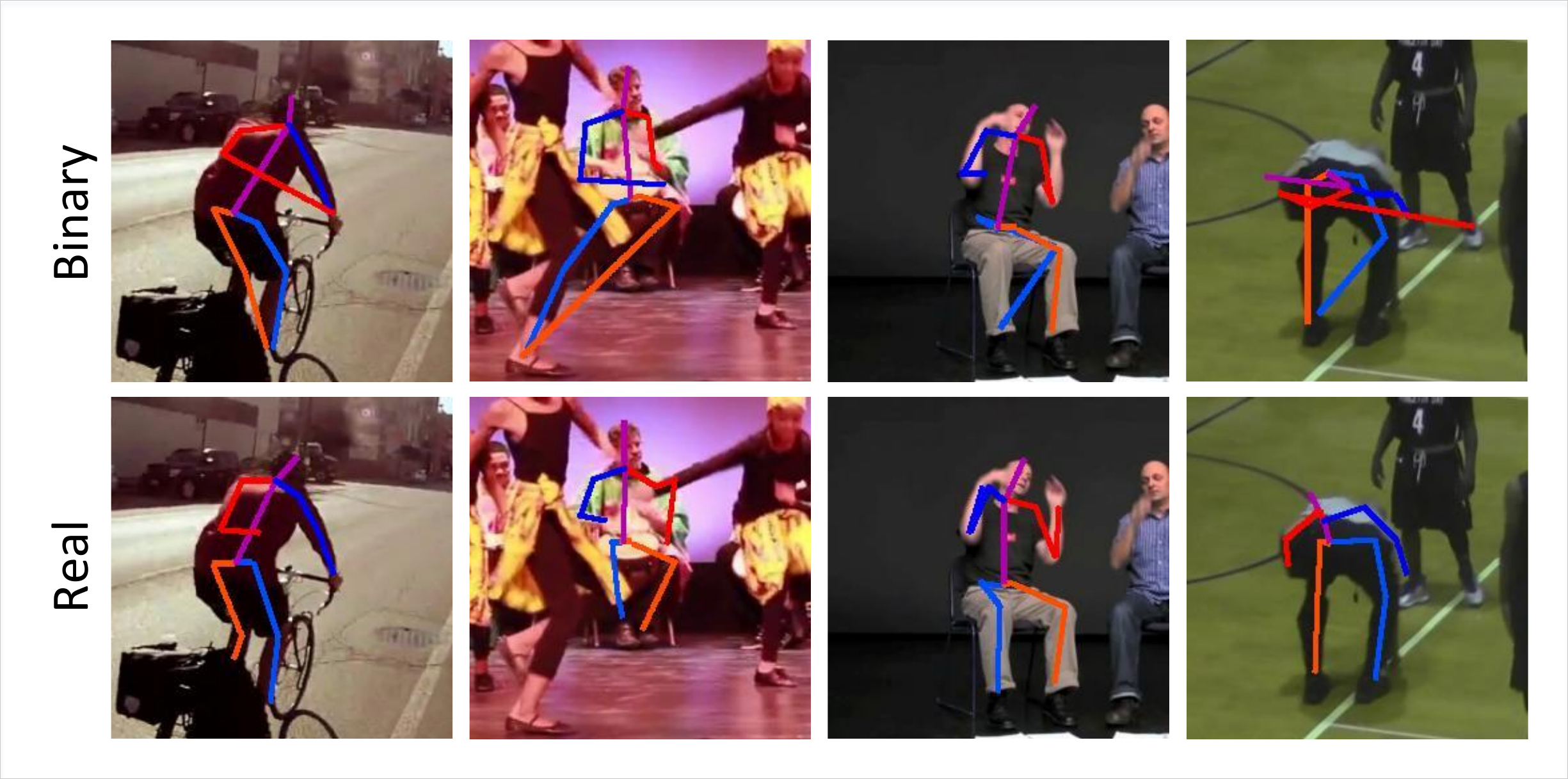}
		\end{center}
		\caption{Examples of failure cases for the binarized HG (first row) and  predictions of its real-valued counterpart (second row). The binary HG misses certain range of poses while having similar accuracy for the correct parts.}
		\label{fig:fails}
        \vspace{-0.3cm}
	\end{figure}

The performance of the original (real-valued) and the binarized HG networks can be seen in Table \ref{tab:bootlneck_vs}. We observe that binarization results in significant performance drop. As we may notice, for almost all parts, there is a large difference in performance which clearly indicates that the binary network has significant less representational power. Some failure cases are shown in Fig.~\ref{fig:fails} illustrating that the binary network was not able to learn some difficult poses. We address this with a better architecture as detailed in the next four Subsections.
    
\begin{table}[!htbp]
    \small
	\begin{center}
		\begin{tabular}{|l|c|c|c|c|c|c|c|} 
			\hline
			Crit. & Bottleneck (real) & Bottleneck (binary)\\
				\hline\hline
				Head  & 94.9 & 90.5 \\
				Shld  & 85.8 & 79.6  \\
				Elbow & 76.9 & 63.0 \\
				Wrist & 71.3 & 57.2 \\
				Hip   & 78.1 & 71.1  \\
				Knee  & 70.1 & 58.2 \\
				Ankle & 63.2 & 53.4 \\
                \hline
				PCKh  & 76.5 &67.2 \\
				\hline
				\# par.  & 3.5M & 3.5M \\
				\hline
			\end{tabular}
		\end{center}
		\caption{PCKh error on MPII dataset for real-valued and binary  bottleneck blocks within the HG network.}
		\label{tab:bootlneck_vs}
\end{table}
    
\subsection{On the Width of Residual Blocks} \label{ssec:wide_block}
    
The original bottleneck block of Fig. \ref{fig:layers_original} is composed of 3 convolutional layers with a filter size of $1\times1$, $3\times3$ and $1\times1$, with the first layer having the role of limiting the width (i.e. the number of channels) of the second layer, thus greatly reducing the number of parameters inside the module. However, it is unclear whether the idea of having a bottleneck structure will be also successful for the binary case, too. Due to the limited representational power of the binary layers, greatly reducing the number of channels  might reduce the amount of information that can be passed from one layer to another, leading to lower performance.
	
	To investigate this, we modify the bottleneck block by increasing the number of channels in the \textit{thin} $3\times3$ layer from 128 to 256. By doing so, we match the number of channels from the first and last layer, effectively removing the ``bottleneck'', and increasing the amount of information that can be passed from one block to another. The resulting \textbf{wider} block is shown in Fig.~\ref{fig:layers_wider}. Here, ``wider''\footnote {The term wider here strictly refers to a ``moderate'' increase in the number of channels in the \textit{thin} layer (up to 256), effectively removing the ``bottleneck''. Except for the naming there is no other resemblance with \cite{zagoruyko2016wide} which performs a study of wide vs deep, using a different building block alongside a much higher number of channels (up to 2048) and without any form of quantization. A similar study falls outside the scope of our work.}  refers to the increased number of channels over the initial \textit{thin} layer.
    
As Table \ref{tab:method_results} illustrates, while this improves performance against the baseline, it also raises the memory requirements. \newline \textbf{Conclusion}: Widening the \textit{thin} layer offers tangible performance improvement, however at a high computational cost.
	
\subsection{On Multi-Scale Filtering} \label{ssec:size_filters}
Small filters have been shown both effective and efficient \cite{simonyan2014very,szegedy2015going} with models being solely made up by a combination of convolutional layers with $3 \times 3$ and/or $1 \times 1$ filters \cite{he2016deep,he2016identity,simonyan2014very}. For the case of real-valued networks, a large number of kernels can be learned. However, for the binary case, the number of possible unique convolutional kernels is limited to $2^k$ states only, where $k$ is the size of the filter. 
	
To address the limited representation power of $3 \times 3$ filters for the binary case, and similarly to \cite{szegedy2017inception}, we largely depart from the block of Fig.~\ref{fig:layers_wider} by proposing the multi-scale structure of Fig.~\ref{fig:layers_mscale}. Note that we implement our multi-scale approach using both larger filter sizes and max-pooling, which greatly increase the effective receptive field within the block. Also, because our goal is to analyze the impact of a multi-scale approach alone, we intentionally keep the number of parameters to a similar level to that of the original bottleneck block of  Fig.~\ref{fig:layers_original}. To this end, we avoid a leap in the number of parameters, by (a) decomposing the $5 \times 5$ filters into two layers of $3 \times 3$  filters, and (b) by preserving the presence of \textit{thin} layer(s) in the middle of the block. 

Given the above, we split the input into two branches. The first (left) branch works at the same scale as the original bottleneck of Fig.~\ref{fig:layers_original} but has a $1 \times 1$ layer that projects the 256 channels into 64 (instead of 128) before going to the $3\times3$ one. The second (right) branch performs a multi-scale analysis by firstly passing the input through a max-pooling layer and then creating two branches, one using a $3 \times 3$ filter and a second one using a $5 \times 5$ decomposed into two $3 \times 3$. By concatenating the outputs of these two sub-branches, we obtain the remaining 64 channels (out of the 128 of the original bottleneck block). Finally, the two main branches are concatenated adding up to 128 channels, which are again back-projected to 256 with the help of a convolutional layer with $1 \times 1$ filters. 

The accuracy of the proposed structure can be found in Table~\ref{tab:method_results}. We can observe a healthy performance improvement at little additional cost and similar computational requirements to the original bottleneck of Fig.~\ref{fig:layers_original}. \newline\textbf{Conclusion}: When designing binarized networks, multi-scale filters should be preferred.
	
\subsection{On $1 \times 1$ Convolutions} \label{ssec:conv1}
	
In the previously proposed block of Fig.~\ref{fig:layers_mscale}, we opted to avoid an increase in the number of parameters, by retaining the two convolutional layers with $1 \times 1$ filters. In this Subsection, by relaxing this restriction, we analyze the influence of $1 \times 1$ filters on the overall network performance.

\begin{figure}[!htb]
		\begin{center}
			\centering
			\includegraphics[height=1.2in,trim={0.1cm 0.1cm 0.1cm 0.1cm},clip]{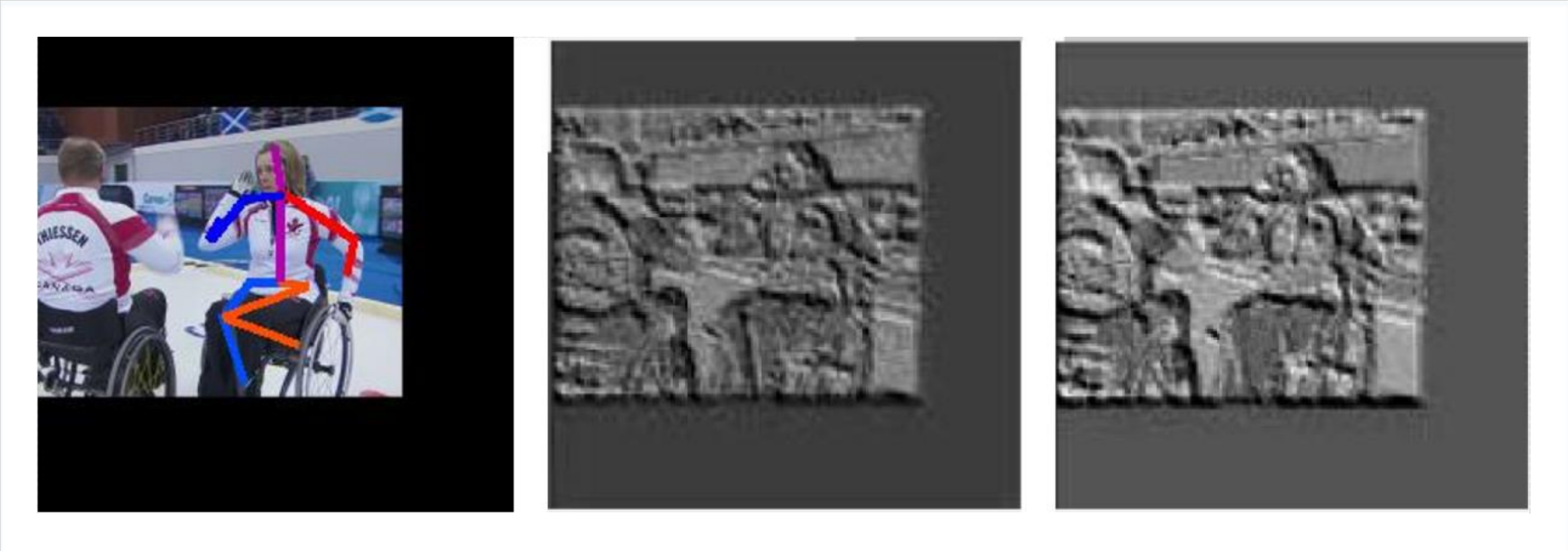}
		\end{center}
		\caption{Examples of features before and after an $1 \times 1$ convolutional layer. Often the features are copied over with little modifications, usually consisting in the details' removal. The contrast was altered for better visualization.}
		\label{fig:ifilters_1x1}
        \vspace{-0.3cm}
	\end{figure}

In particular, we remove all convolutional layers with $1 \times 1$ filters from the multi-scale block of Fig.~\ref{fig:layers_mscale}, leading to the structure of  Fig.~\ref{fig:layers_no1x1}. Our motivation to remove $1 \times 1$ convolutions for the binary case is the following: because $1 \times 1$ filters are limited to two states only (either 1 or -1) they have a very limited learning power. Due to their nature, they behave as simple filters deciding when a certain value should be passed or not. In practice, this allows the input to pass through the layer with little modifications, sometimes actually blocking ``good features'' and hurting the overall performance by a noticeable amount. This is particularly problematic for the task of landmark localization, where a high level of detail is required for successful localization. Examples of this problem are shown in Fig.~\ref{fig:ifilters_1x1}.

Results reported in Table~\ref{tab:method_results} show that by removing $1 \times 1$ convolutions, performance over the baseline is increased by more than 8\%. Even more interestingly, the newly introduced block outperforms the one of Subsection~\ref{ssec:wide_block}, while having less parameters, which shows that the presence of $1 \times 1$ filters limits the performance of binarized CNNs. \newline \textbf{Conclusion}: The use of $1 \times 1$ convolutional filters on binarized CNNs has a detrimental effect on performance and should be avoided.
	
\subsection{On Hierarchical, Parallel \& Multi-Scale} \label{ssec:better}
	
\begin{figure*}[!htb]
		\centering
		\begin{subfigure}[t]{0.3\textwidth}
			\centering
			\includegraphics[height=2in,trim={0.5cm 0.5cm 0.5cm 0.5cm},clip]{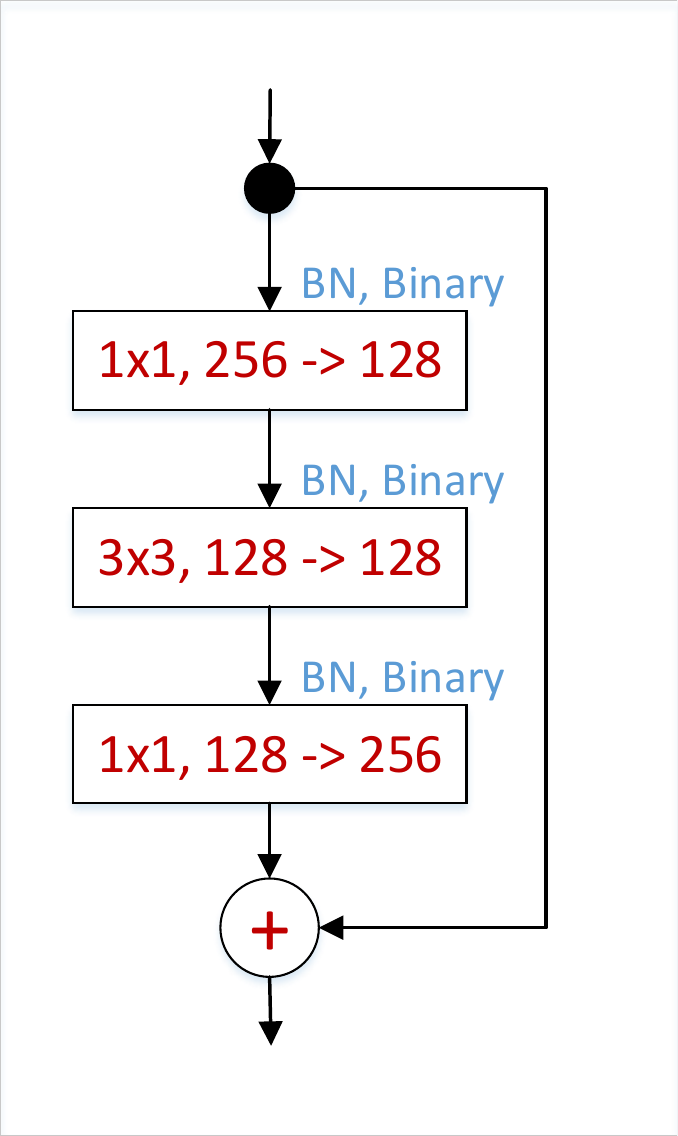}
			\caption{The \textbf{Original Bottleneck} block with pre-activation, as defined in \cite{he2016identity}. Its binarized version is described in Section~\ref{ssec:naive}.}
			\label{fig:layers_original}
		\end{subfigure}%
		~ 
		\begin{subfigure}[t]{0.3\textwidth}
			\centering
			\includegraphics[height=2in,trim={0.5cm 0.5cm 0.5cm 0.5cm},clip]{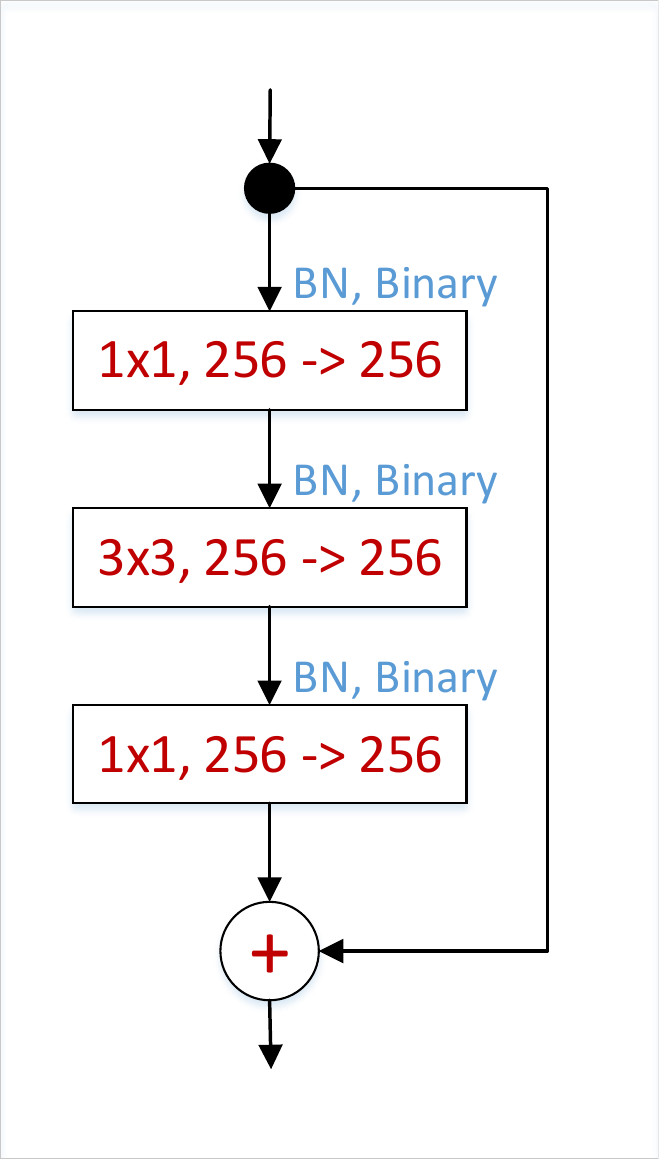}
			\caption{The \textbf{Wider} version of (a) produced by increasing the number of filters in the second layer. See Subsection~\ref{ssec:wide_block}. }
			\label{fig:layers_wider}
		\end{subfigure}
		~ 
		\begin{subfigure}[t]{0.3\textwidth}
			\centering
			\includegraphics[height=2in,trim={0.5cm 0.5cm 0.5cm 0.5cm},clip]{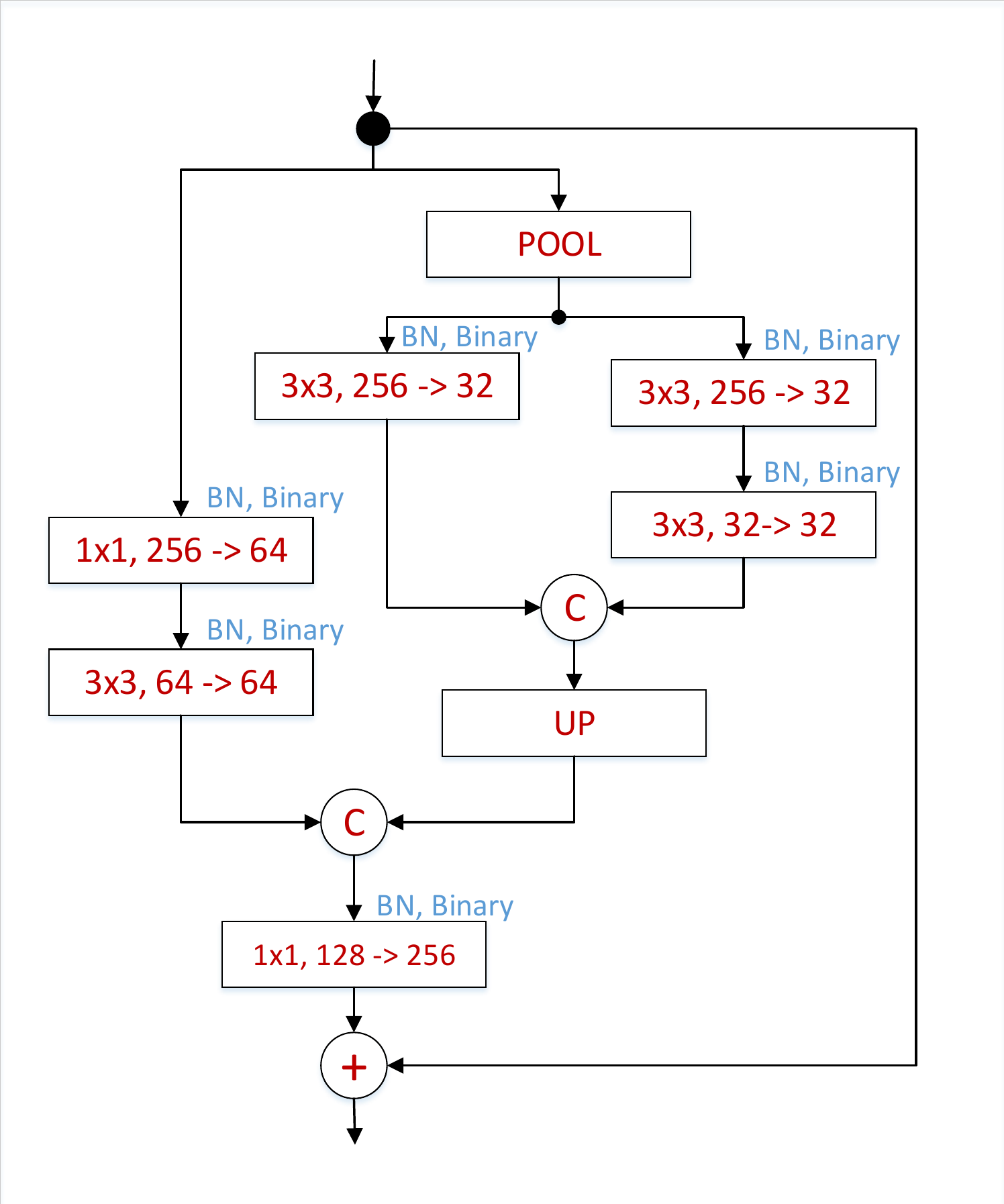}
			\caption{Largely departing from (b), this block consists of \textbf{Multi-Scale (MS)} filters for analyzing the input at multiple scales. See Subsection~\ref{ssec:size_filters}.}
			\label{fig:layers_mscale}
		\end{subfigure}
		~ 
		\begin{subfigure}[t]{0.45\textwidth}
			\centering
			\includegraphics[height=2in,trim={0.5cm 0.5cm 0.5cm 0.5cm},clip]{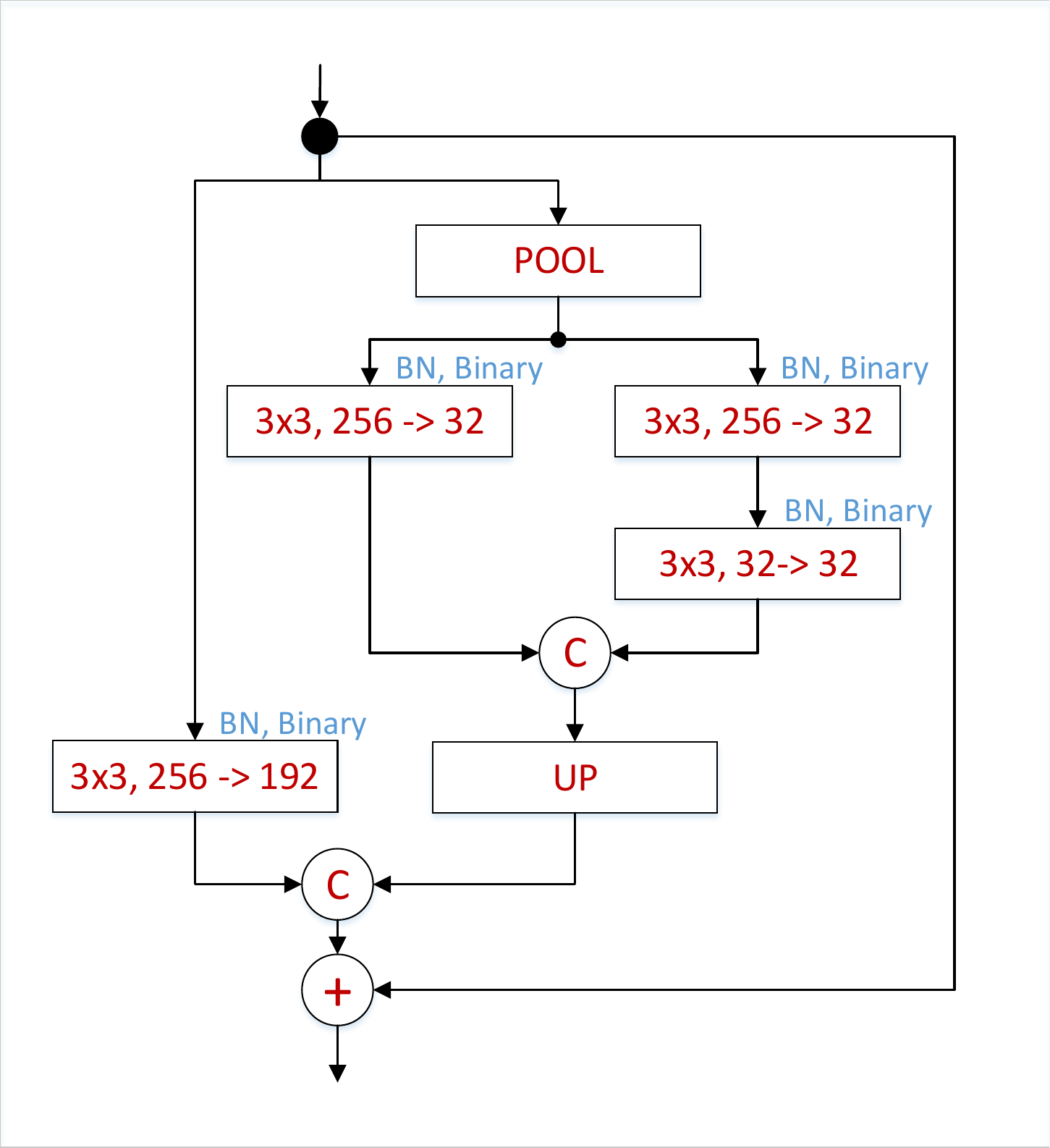}
			\caption{A variant of the MS block introduced in (c) after removing all convolutional layers with $1\times 1$ filters (\textbf{MS Without $1\times 1$ filters}). See  Subsection~\ref{ssec:size_filters}.}
			\label{fig:layers_no1x1}
		\end{subfigure}
		~ 
		\begin{subfigure}[t]{0.45\textwidth}
			\centering
			\includegraphics[height=2in,trim={0.5cm 0.5cm 0.5cm 0.5cm},clip]{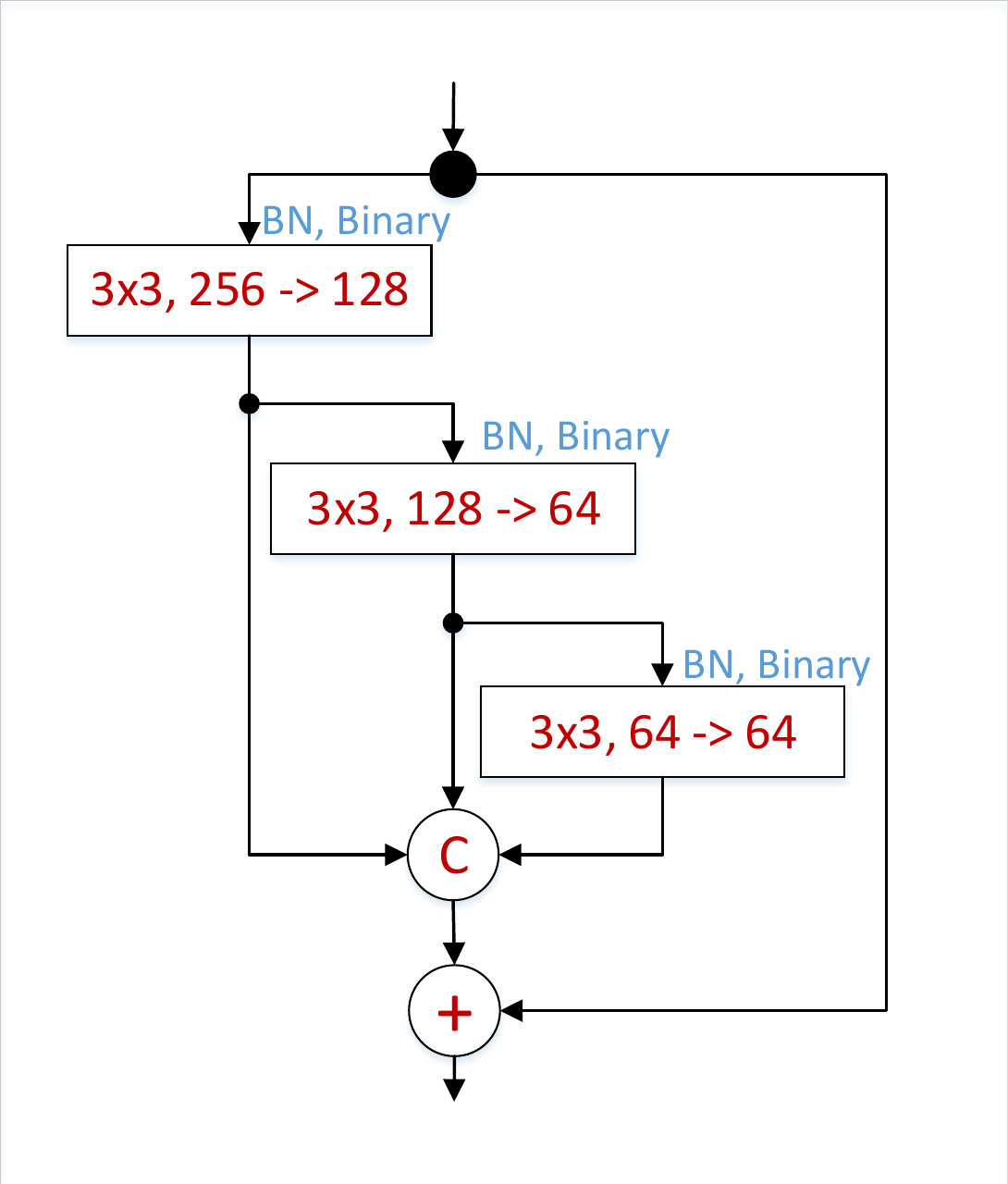}
			\caption{The proposed \textbf{Hierarchical, Parallel \& MS} (denoted in the paper as \textbf{Ours, final}) block incorporates all ideas from (b), (c) and (d) with an improved gradient flow. See  Subsection~\ref{ssec:better}}
			\label{fig:layers_parallel}
		\end{subfigure}
        \\
        \vspace{0.5cm}
		\caption{Different types of blocks described and evaluated. Our best performing block is shown in figure (e). A layer is depicted as a rectangular block containing: its filter size, number of input channels and the number of output channels). ``C" - denotes concatenation operation and ``+" an element-wise sum.}
		\label{fig:layers}
	\end{figure*}

Binary networks are even more sensitive to the problem of fading gradients \cite{courbariaux2016binarized,rastegari2016xnor}, and for our network we found that the gradients are up to 10 times smaller than those corresponding to its real-valued counterpart. To alleviate this, we design a new module which has the form of a hierarchical, parallel multi-scale structure allowing, for each resolution, the gradients to have 2 different paths to follow, the shortest of them being always 1. The proposed block is depicted in Fig.~\ref{fig:layers_parallel}. Note that, in addition to better gradient flow, our design encompasses all the findings from the previous Subsections: (a) no convolutional layers with $1 \times 1$ filters should be used, (b) the block should preserve its width as much as possible (avoiding large drops in the number of channels), and (c) multi-scale filters should be used.

Contrary to the blocks described in Subsections~\ref{ssec:wide_block} -~\ref{ssec:conv1}, where the gradients may need to pass through two more layers before reaching the output of the block, in the newly proposed module, each convolutional layer has a direct path that links it to the output, so that at any given time and for all the layers within the module the shortest possible path is equal to 1. The presence of a hierarchical structure inside the module efficiently accommodates larger filters (up to $7 \times 7$), decomposed into convolutional layers with $3 \times 3$ filters. Furthermore, our design avoids the usage of an element-wise summation layer as for example in \cite{xie2016aggregated,szegedy2017inception}, further improving the gradient flow and keeping the complexity under control.

As we can see in Table~\ref{tab:method_results}, the proposed block matches and even outperforms the block proposed in Section~\ref{ssec:size_filters} having far less parameters.
	\begin{table}[!htbp]
    \small
		\begin{center}
			\begin{tabular}{|l|c|c|}
				\hline
				Block type & \# params & PCKh \\
				\hline\hline
				Bottleneck (original) (Fig.~\ref{fig:layers_original})& 3.5M & 67.2\%\\
				Wider (Fig.~\ref{fig:layers_wider})& 11.3M & 70.7\% \\
				Multi-Scale (MS) (Fig.~\ref{fig:layers_mscale})& 4.0M & 69.3\%\\
				MS without 1x1 filters (Fig.~\ref{fig:layers_no1x1})& 9.3M & 75.5\%\\
                \hline
                {\begin{tabular}{c}
\textbf{Hierarchical, Parallel \& MS } \\ \textbf{(Ours, Final)} (Fig.~\ref{fig:layers_parallel})\end{tabular}}

				& \textbf{6.2M} & \textbf{76\%}\\
				\hline
			\end{tabular}
		\end{center}
		\caption{PCKh-based comparison of different blocks on MPII validation set. \# params refers to the number of parameters of the whole network.}
		\label{tab:method_results}
	\end{table}
\newline \textbf{Conclusion}: Good gradient flow and hierarchical multi-scale filtering are crucial for high performance without excessive increase in the parameters of the binarized network.

\section{Proposed vs Bottleneck} \label{sec:binaryvsreal}
In this Section, we attempt to make a fair comparison between the performance of the proposed block (\textbf{Ours, Final}, as in Fig.~\ref{fig:layers_parallel}) against that of the original bottleneck module (Fig.~\ref{fig:layers_original}) by taking two important factors into account:
\begin{itemize}[leftmargin=*]
\setlength\itemsep{-0.35em}
\item
Both blocks should have the same number of parameters.
\item
The two blocks should be compared for the case of binary but also real-valued networks.  
\end{itemize}
With this in mind, in the following Sections, we show that:
\begin{itemize}[leftmargin=*]
\setlength\itemsep{-0.35em}
\item
The proposed block largely outperforms a bottleneck with the same number of parameters for the binary case. 
\item
The proposed block also outperforms a bottleneck with the same number of parameters for the real case but in this case the performance difference is smaller.   
\end{itemize}
We conclude that, for the real case, increasing the number of parameters (by increasing width) results in performance increase; however this is not the case for binary networks where a tailored design as the one proposed here is needed.
    
  \begin{table}[!htbp]
            \small
		\begin{center}
			\begin{tabular}{|l|c|c|}
				\hline
				Layer type & \# parameters & PCKh \\
				\hline\hline
                Bottleneck (Original) (Fig.~\ref{fig:layers_original}) & 3.5M & 67.2\%\\
                \hline
				Wider (Fig.~\ref{fig:layers_wider}) & 11.3M & 70.7\%\\
\hline
Bottleneck (wider) + no $1 \times 1$ & 5.8M & 69.5\%\\
\hline
\textbf{(Ours, Final)} (Fig.~\ref{fig:layers_parallel}) & \textbf{6.2M} & \textbf{76\%}\\			
				\hline
			\end{tabular}
		\end{center}
		\caption{PCKh-based performance on MPII validation set for binary blocks: the \# parameters of the original bottleneck are increased to match the \# parameters of the proposed block. This firstly gives rise to the Wider block and its variant without the $1 \times 1$ Convolutions. }
		\label{tab:binary_comp1}
        \vspace{-0.5cm}
	\end{table}
    
\subsection{Binary} \label{sec:binary}

To match the number of parameters between the proposed and bottleneck block, we follow two paths. Firstly, we increase the number of parameters of the bottleneck: (a) a first way to do this is to make the block wider as described in Section \ref{ssec:wide_block}. Note that in order to keep the number or input-output channels equal to 256, the resulting block of Fig. \ref{fig:layers_wider} has a far higher number of parameters than the proposed block. Despite this, the performance gain is only moderate (see Section \ref{ssec:wide_block} and Table~\ref{tab:binary_comp1}). (b) Because we found that the $1 \times 1$ convolutional layers have detrimental effect to the performance of the Multi-Scale block of Fig.~\ref{fig:layers_mscale}, we opted to remove them from the bottleneck block, too. To this end, we modified the Wider module by (a) removing the $1 \times 1$ convolutions and (b) halving the number of parameters in order to match the number of parameters of the proposed  block. The results in Table~\ref{tab:binary_comp1} clearly show that this modification is helpful but far from being close to the performance achieved by the proposed block. 
 
 \begin{table}[!htbp]
              \small
		\begin{center}
			\begin{tabular}{|l|c|c|}
				\hline
				Layer type & \# parameters & PCKh \\
				\hline \hline
                Bottleneck (original) & 3.5M & 67.2\%\\
\hline
\textbf{(Ours, Final)} (Fig.~\ref{fig:layers_wider_intro}) & \textbf{4.0M} & \textbf{72.7\%}\\
				\hline
			\end{tabular}
		\end{center}
		\caption{PCKh-based performance on MPII validation set for binary blocks: the \# parameters of the proposed block are decreased to match the \# parameters of the bottleneck.} 
        \vspace{-0.55cm} 
		\label{tab:binary_comp2}
	\end{table}
   
Secondly, we decrease the number of parameters in the proposed block to match the number of parameters of the original bottleneck. This block is shown in Fig. \ref{fig:layers_wider_intro}. To this end, we reduced the number of input-output channels of the proposed block from 256 to 192 so that the number of channels in the first layer are modified from [256 $\rightarrow$ 128, $3\times3$] to [192$\rightarrow$96, $3\times3$], in the second layer from [128$\rightarrow$64, $3\times3$] to [96$\rightarrow$48, $3\times3$] and in the third layer from [64$\rightarrow$64, $3\times3$] to [48$\rightarrow$48, $3\times3$].  Notice, that even in this case, the proposed binarized module outperforms the original bottleneck block by more than 5\% (in absolute terms) while both have very similar number of parameters (see Table~\ref{tab:binary_comp2}).  	
    
\subsection{Real} \label{sec:real}
\vspace{-0.1cm}
While the proposed block was derived from a binary perspective, Table~\ref{tab:real} shows that a significant performance gain is also observed for the case of real-valued networks. In order to quantify this performance improvement and to allow for a fair comparison, we increase the number of channels inside the original bottleneck block so that both networks have the same depth and a similar number of parameters. Even in this case, our block outperforms the original block although the gain is smaller than that observed for the binary case. We conclude that for real-valued networks performance increase can be more easily obtained by simply increasing the number of parameters, but for the binary case a better design is needed as proposed in this work. 
\newline

    	\begin{table}[!htbp]
        \small
		\begin{center}
			\begin{tabular}{|l|c|c|}
				\hline
				Layer type & \# parameters & PCKh \\
				\hline\hline
				Bottleneck (wider) & 7.0M & 83.1\%\\
\hline
				\textbf{(Ours, Final)} & \textbf{6.2M} & \textbf{85.5\%}\\
				\hline
			\end{tabular}
		\end{center}
		\caption{PCKh-based performance on MPII validation set for real-valued blocks: Our block is compared with a wider version of the original bottleneck so that both blocks have similar \# parameters.}
		\label{tab:real}
        \vspace{-0.3cm}
	\end{table}
    
	\section{Ablation studies} \label{sec:ablation}
    
	In this Section, we present a series of other architectural variations and their effect on the performance of our binary network. All reported results are obtained using the proposed block of Fig.~\ref{fig:layers_parallel} coined \textbf{Ours, Final}. We focus on the effect of augmentation and different losses which are novel experiments not reported in \cite{rastegari2016xnor}, and then comment on the effect of pooling, ReLUs and performance speed-up providing more details in the supplementary material. \newline
\textbf{Is Augmentation required?} Recent works have suggested that binarization is an extreme case of regularization \cite{courbariaux2015binaryconnect,courbariaux2016binarized,merolla2016deep}. In light of this, one might wonder whether data augmentation is still required.  Table~\ref{tab:aug_results} shows that in order to accommodate the presence of new poses and/or scale variations, data augmentation is very helpful providing a large increase (4\%) in performance.
\begin{table}[!htbp]
\small
		\begin{center}
			\begin{tabular}{|l|c|c|}
				\hline
				Layer type & \# parameters & PCKh \\
				\hline\hline
				(Ours, Final) (No Aug.) & 6.2M & 72.1\%\\
\hline
				\textbf{(Ours, Final) + Aug.} & \textbf{6.M} & \textbf{76\%}\\
				\hline
			\end{tabular}
		\end{center}
		\caption{The effect of using augmentation when training our binary network in terms of PCKh-based performance on MPII validation set.}
		\label{tab:aug_results}
	\end{table}
\newline \textbf{The effect of loss}. We trained our binary network to predict a set of heatmaps, one for each landmark \cite{tompson2014joint}. To this end, we experimented with two types of losses: the first one places a Gaussian around the correct location of each landmark and trains using a pixel-wise L2 loss \cite{tompson2014joint}. However, the gradients generated by this loss are usually small even for the case of a real-valued network. Because binarized networks tend to amplify this problem, as an alternative, we also experimented with the Sigmoid cross-entropy pixel-wise loss typically used for detection tasks \cite{zhang2015fine}. We found that the use of the Sigmoid cross-entropy pixel-wise loss increased the gradients by 10-15x (when compared to the L2 loss), offering a 2\% improvement (see Table~\ref{tab:loss_results}), after being trained for the same number of epochs.
\begin{table}[!htbp]
\small
		\begin{center}
			\begin{tabular}{|l|c|c|}
				\hline
				Layer type & \# parameters & PCKh \\
				\hline\hline
				(Ours, Final) + L2  & 6.2M & 73.8\%\\
                \hline
				 \textbf{(Ours, Final) + Sigmoid} & \textbf{6.2M} & \textbf{76\%}\\
				\hline
			\end{tabular}
		\end{center}
		\caption{The effect of using different losses (Sigmoid vs L2) when training our binary network in terms of PCKh-based performance on MPII validation set.}
		\label{tab:loss_results}
\end{table}
\newline\textbf{Pooling type.} In line with \cite{rastegari2016xnor}, we found that max-pooling outperforms average pooling, resulting in 4\% performance increase. See also supplementary material. \newline
\textbf{ReLUs.} In line with \cite{rastegari2016xnor}, we found that by adding a ReLU activation after each convolutional layer performance is increased, observing a 2\% performance improvement. See also supplementary material. \newline
\textbf{Performance.} In line with \cite{rastegari2016xnor}, we observed speedups of up to 3.5x when compared against cuBLAS. We did not conduct experiments on CPUs. However, since we used the same method for binarization as in \cite{rastegari2016xnor}, speed improvements of the order of 58x, are to be expected allowing the system to run in real-time on a CPU using a single core. In terms of memory compression, we can achieve a compression rate of 39x when compared against its single precision counterpart from Torch. See also supplementary material.

\section{Comparison with state-of-the-art} \label{sec:results}

In this Section, we compare our method against the current state-of-the-art for human pose estimation and 3D face alignment. Our final system comprises a single HG network but replaces the real-valued bottleneck block used in \cite{newell2016stacked} with the proposed binary, parallel, multi-scale block trained with the improvements detailed in Section~\ref{sec:ablation}. Moreover, to show that the proposed block generalizes well producing consistent results across various datasets and tasks, our supplementary material provides the results of a facial part segmentation experiment.
\newline \textbf{Human Pose Estimation.} As in all previous experiments, we used the standard training-validation partition of MPII \cite{bulat2016human,newell2016stacked}. In Table~\ref{tab:mpii_results}, we report the performance of (a) the proposed binary block, (b) the proposed block when implemented and trained with real values, (c) the real-valued stacked HG network consisting of 8 stacked single real-valued HG networks trained with intermediate supervision (state-of-the-art on MPII {\cite{newell2016stacked}}) and, finally, (d) the same real-valued network as in (c) where the bottleneck block is replaced by our proposed block.

The results are shown in Table~\ref{tab:mpii_results}. We observe that when a single HG network with the proposed block is trained with real weights, its performance reaches that of \cite{newell2016stacked}. This result clearly illustrates the enhanced learning capacity of the proposed block. Moreover, there is still a gap between the binary and real-valued version of the proposed block indicating that margin for further improvement is possible. We also observe that a full-sized model (with 8 HG networks) based on the proposed block performs slightly better than the original network from {\cite{newell2016stacked}}, indicating that, for the real-valued case, the new block is more effective than the original one when a smaller computational budget is used. 

	\begin{table}[!htbp]
    \small
	\begin{center}
		\begin{tabular}{|l|c|c|c|c|c|c|c|} 
			\hline
			Crit. & \cite{newell2016stacked} & Ours, bin. & Ours[1x], real & Ours[8x], real\\
				\hline\hline
				Head  & 97.3 & 94.7 & 96.8 & 97.4\\
				Shld  & 96.0 & 89.6 & 93.8 & 96.0 \\
				Elbow & 90.2 & 78.8 & 86.4 & 90.7\\
				Wrist & 85.2 & 71.5 & 80.3 & 86.2\\
				Hip   & 89.1 & 79.1 & 87.0 & 89.6 \\
				Knee  & 85.1 & 70.5 & 80.4 & 86.1\\
				Ankle & 82.0 & 64.0 & 75.7 & 83.2\\
                \hline
				PCKh  & 89.3 &78.1 & 85.5 & 89.8\\
				\hline
				\# par.  & 25M & 6M & 6M & 25M\\
				\hline
			\end{tabular}
		\end{center}
		\caption{PCKh-based comparison on MPII validation set.}
		\label{tab:mpii_results}
	\end{table}

\textbf{Face alignment.} We used three very challenging datasets for large pose face alignment, namely AFLW {\cite{kostinger2011annotated}}, AFLW-PIFA {\cite{jourabloo2015pose}}, and AFLW2000-3D {\cite{zhu2016face}}. The evaluation metric is the Normalized Mean Error (NME) {\cite{jourabloo2015pose}.}

AFLW is a large-scale face alignment dataset consisting of 25993 faces annotated with up to 21 landmarks. The images are captured in arbitrary conditions exhibiting a large variety of poses and expressions. As Table{~\ref{tab:aflw_full_results}} shows, our binarized network outperforms the current state-of-the-art methods, all of which use large real-valued CNNs.

	\begin{table}[!htbp]
    \small
	\begin{center}
		\begin{tabular}{|l|c|c|c|c|} 
			\hline
			Method & [0,30] & [30,60] & [60,90] & mean\\
				\hline\hline
				HyperFace \cite{ranjan2016hyperface} & 3.93 & 4.14 & 4.71 & 4.26\\
                \hline
				AIO \cite{ranjan2017all} & 2.84 & 2.94 & 3.09 &2.96\\
                \hline
				\textbf{Ours}   & \textbf{2.77} & \textbf{2.86} & \textbf{2.90} & \textbf{2.85} \\
				\hline
			\end{tabular}
		\end{center}
		\caption{NME-based (\%) comparison on AFLW test set. The evaluation is done on the test set used in \cite{ranjan2017all}. }
		\label{tab:aflw_full_results}
	\end{table}
    
\begin{figure}[!htb]\hspace*{0.3cm}
             \begin{subfigure}[t]{0.2\textwidth}\hspace*{-0.2cm}
             \begin{center}
             \centering
			\includegraphics[height=1.1in,trim={0.5cm 0.5cm 0.5cm 0.5cm},clip]{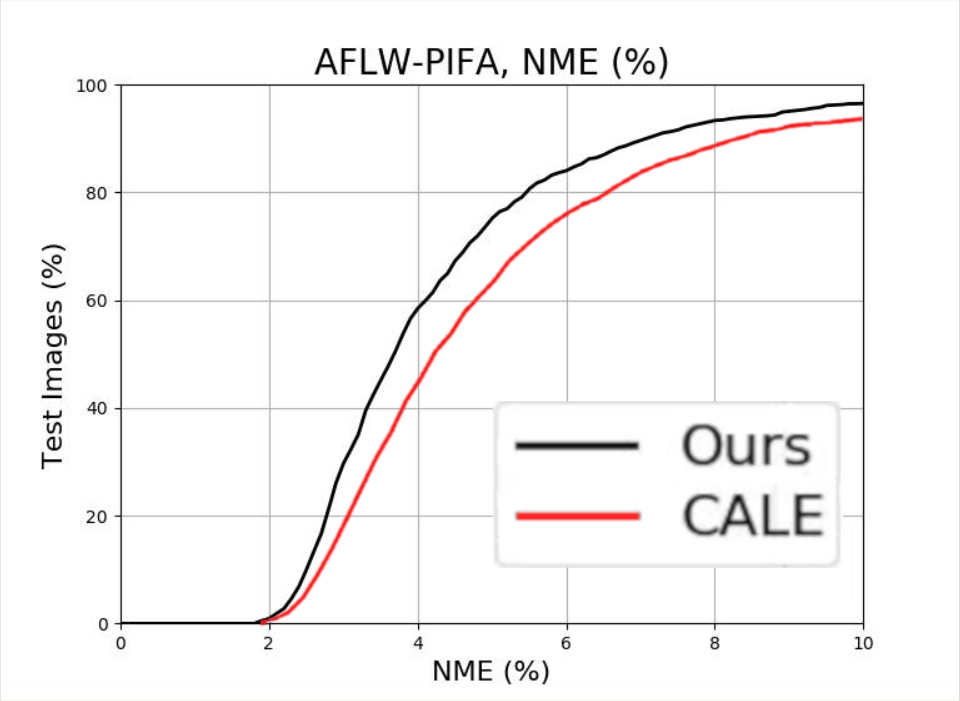} 
            \caption{}
            \label{fig:pifa_a}
            \end{center}
            \end{subfigure}
            ~
             \begin{subfigure}[t]{0.2\textwidth}\hspace*{-0.25cm}
             \begin{center}
             \centering
            \includegraphics[height=1.1in,trim={0cm 0.5cm 0cm 0.5cm},clip]{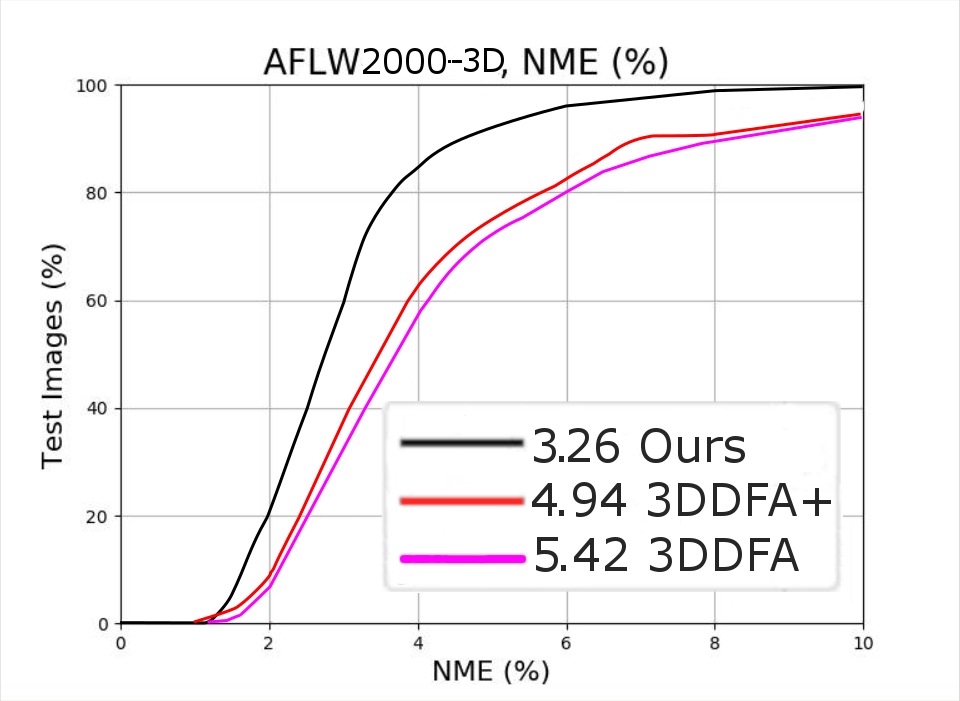}
            \caption{}
            \label{fig:pifa_b}
            \end{center}
            \end{subfigure}
		\caption{Cumulative error curves (a) on AFLW-PIFA, evaluated on all 34 points (CALE is the method of \cite{bulat2016convolutional}), (b) on AFLW2000-3D on all points computed on a random subset of 696 images equally represented in $[0^\circ,30^\circ], [30^\circ,60^\circ], [60^\circ,90^\circ]$ (see also \cite{zhu2016face}).}
		\label{fig:pifa}
	\end{figure}

AFLW-PIFA \cite{jourabloo2015pose} is a grey-scale subset of AFLW \cite{kostinger2011annotated}, consisting of 5200 images (3901 for training and 1299 for testing) selected so that there is a balanced number of images for yaw angle in $[0^\circ,30^\circ], [30^\circ,60^\circ] \textrm{ and } [60^\circ,90^\circ]$. All images are annotated with 34 points from a 3D perspective. Fig.~\ref{fig:pifa_a} shows our results on AFLW-PIFA. When evaluated on both visible and occluded points, our method improves upon the current best result of \cite{bulat2016convolutional} (which uses real weights) by more than 10\%. For additional numerical results on AFLW-PIFA, see also our supplementary material.
	
AFLW2000-3D is a subset of AFLW re-annotated by \cite{zhu2016face} from a 3D perspective with 68 points. We used this dataset only for evaluation. The training was done using the first 40000 images from 300W-LP \cite{zhu2016face}. As Fig.~\ref{fig:pifa_b} shows, on AFLW2000-3D, the improvement over the  state-of-the-art method of \cite{zhu2016face} (real-valued) is even larger. As further results in our supplementary material show, while our method improves over the entire range of poses, the gain is noticeably higher for large poses ($[60^\circ-90^\circ]$) where we outperform \cite{zhu2016face} by more than 40\%. 
\vspace{-0.75cm}

\textbf{Training.}  All models were trained from scratch following the algorithm described in \cite{rastegari2016xnor} and using rmsprop \cite{tieleman2012lecture}. The initialization was done as in \cite{he2016deep}. For human pose estimation, we randomly augmented the data with rotation (between -$40^o$ and $40^o$ degrees), flipping and scale jittering (between 0.7 and 1.3). We  trained the network for 100 epochs, dropping the learning rate four times, from 2.5e-4 to 5e-5. A similar procedure was applied to the models for 3D face alignment, with the difference that the training was done for 55 epochs only. The input was normalized between 0 and 1 and all described networks were trained using the binary cross-entropy loss. The models were implemented with Torch7 \cite{collobert2011torch7}. 
	
	\section{Conclusion} \label{sec:conclusion}
	We proposed a novel block architecture, particularly tailored for binarized CNNs for the tasks of human pose estimation and face alignment.  During the process, we exhaustively evaluated various design choices, identified performance bottlenecks and proposed solutions. We showed that our hierarchical, parallel and multi-scale block enhances representational power, allowing for stronger relations to be learned without excessively increasing the number of network parameters. The proposed architecture is efficient and can run on limited resources.
    
    \section{Acknowledgment} Adrian Bulat was funded by a PhD scholarship from the University of Nottingham. Georgios Tzimiropoulos was supported in part by the EPSRC project EP/M02153X/1 Facial Deformable Models of Animals.

\pagebreak
	
	{\small
		\bibliographystyle{ieee}
		\bibliography{egbib}
	}

\appendix
\renewcommand{\thesection}{A\arabic{section}}

\section{Additional ablation studies} \label{sec:app}

This section provides additional details for some of the ablation studies reported in Section 6. \newline \textbf{Pooling type.} In the context of binary networks, and because the output is restricted to 1 and -1, max-pooling  might result in outputs full of 1s only. To limit this effect, we placed the activation function before the convolutional layers as proposed in \cite{he2016identity,rastegari2016xnor}. Additionally, we opted to replace max-pooling with average pooling. However, this leads to slightly worse results (see Table~\ref{tab:pool_results}). In practice, we found that the use of blocks with pre-activation suffices and that the ratio of 1 and -1 is close to 50\% even after max-pooling.
\begin{table}[!htbp]
	\begin{center}
		\begin{tabular}{|l|c|c|}
			\hline
			Layer type & \# parameters & PCKh \\
			\hline\hline
			(Ours, Final) + Average  & 6.2M & 71.9\%\\
			\textbf{(Ours, Final) + Max}  & \textbf{6.2M} & \textbf{76\%}\\
			\hline
		\end{tabular}
	\end{center}
	\caption{The effect of using different pooling methods when training our binary network in terms of PCKh-based performance on MPII validation set.}
	\label{tab:pool_results}
\end{table}
\newline \textbf{With or without ReLU.} Because during the binarization process all ReLU layers are replaced with the Sign function, one might wonder if ReLUs are still useful for the binary case. Our findings are in line with the ones reported in \cite{rastegari2016xnor}. By adding a ReLU activation after each convolutional layer, we observe a 2\% performance improvement (see Table~\ref{tab:method_results_relu}), which can be attributed to the added non-linearity, particularly useful for training very deep architectures.
\begin{table}[!htbp]
	\begin{center}
		\begin{tabular}{|l|c|c|}
			\hline
			Layer type & \# parameters & PCKh \\
			\hline\hline
			(Ours, Final) & 6.2M & 76\%\\
			\hline
			\textbf{(Ours, Final) + ReLU} & \textbf{6.2M} & \textbf{77.8\%}\\
			\hline
		\end{tabular}
	\end{center}
	\caption{The effect of using ReLU when training our binary network in terms of PCKh-based performance on MPII validation set.}
	\label{tab:method_results_relu}
\end{table}
\newline \textbf{Performance.} In theory, by replacing all floating-point multiplications with bitwise XOR and making use of the SWAR (Single instruction, multiple data within a register) \cite{rastegari2016xnor,courbariaux2016binarized}, the number of operations can be reduced up to 32x when compared against the multiplication-based convolution. However, in our tests, we observed speedups of up to 3.5x, when compared against cuBLAS, for matrix multiplications, a result being in accordance with those reported in \cite{courbariaux2016binarized}. As GPUs are already available on mobile devices, we did not conduct experiments on CPUs. However, given the fact that we used the same method for binarization as in \cite{rastegari2016xnor}, similar improvements in terms of speed, of the order of 58x, are to be expected: as the real-valued network takes 0.67 seconds to do a forward pass on a i7-3820 using a single core, a speedup close to x58 will allow the system to run in real-time. 

\begin{figure}[!htb]
	\begin{center}
		\centering
		\includegraphics[height=1.8in,trim={0cm 0cm 0cm 0cm},clip]{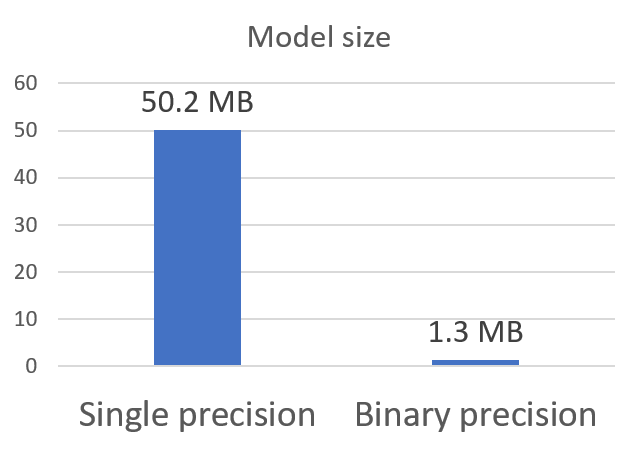}
	\end{center}
	\vspace{-0.2cm}
	\caption{Memory compression ratio. By binarizing the weights and removing the biases, we achieve a compression rate of 39x when compared against the single precision model.}
	\label{fig:memory}
\end{figure}

In terms of memory compression, by removing the biases, which have minimum impact (or no impact at all) on performance, and by grouping and storing every 32 weights in one variable, we can achieve a compression rate of 39x when compared against the single precision counterpart of Torch. See also Fig.~\ref{fig:memory}.
\vspace{-0.25cm}
\vspace{-0.5cm}
\section{Additional face alignment results}

This section provides additional numerical results on AFLW-PIFA and AFLW2000-3D.

\begin{table}[!htbp]
	\begin{center}
		\begin{tabular}{|l|c|c|c|c|}
			\hline
			PIFA \cite{jourabloo2015pose}& RCPR \cite{burgos2013robust} & PAWF \cite{jourabloo2016large}& CALE \cite{bulat2016convolutional}& Ours \\
			\hline\hline
			8.04  & 6.26 & 4.72 & \textbf{2.96} & 3.02\\
			\hline
		\end{tabular}
	\end{center}
	\caption{NME-based (\%) comparison on AFLW-PIFA evaluated on visible landmarks only. The results for PIFA, RCPR and PAWF are taken from \cite{jourabloo2016large}.}
	\label{tab:pifaall}
\end{table}

\begin{table}[!htbp]
	\begin{center}
		\begin{tabular}{|l|c|}
			\hline
			CALE \cite{bulat2016convolutional}& Ours \\
			\hline\hline
			4.97 & \textbf{4.47}\\
			\hline
		\end{tabular}
	\end{center}
	\caption{NME-based (\%) based comparison on AFLW-PIFA evaluated on all 34 points, both visible and occluded. }
	\label{tab:pifa}
\end{table}

\begin{table}[!htbp]
	\begin{center}
		\begin{tabular}{|l|c|c|c|c|}
			\hline
			Method & [0,30] & [30,60] & [60,90] & Mean  \\
			\hline\hline
			RCPR(300W) \cite{burgos2013robust}& 4.16 & 9.88 & 22.58 & 12.21  \\
			RCPR(300W-LP) \cite{burgos2013robust}& 4.26 & 5.96 & 13.18 & 7.80  \\
			ESR(300W) \cite{cao2014face}& 4.38 & 10.47 & 20.31 & 11.72  \\
			ESR(300W-LP) \cite{cao2014face} & 4.60 & 6.70 & 12.67 & 7.99  \\
			SDM(300W) \cite{xiong2013supervised}& 3.56 & 7.08 & 17.48 & 9.37  \\
			SDM(300W-LP) \cite{xiong2013supervised}& 3.67 & 4.94 & 9.76 & 6.12  \\
			3DDFA \cite{zhu2016face} & 3.78 & 4.54 & 7.93 & 5.42 \\
			3DDFA+SDM \cite{zhu2016face} & 3.43 & 4.24 & 7.17 & 4.94  \\
			\hline
			\textbf{Ours} & \textbf{2.47} & \textbf{3.01} & \textbf{4.31} & \textbf{3.26}\\
			\hline
		\end{tabular}
	\end{center}
	\caption{NME-based (\%) based comparison on AFLW2000-3D evaluated on all 68 points, both visible and occluded. The results for RCPR, ESR and SDM are taken from \cite{zhu2016face}. }
	\label{tab:compreAFLW2000}
\end{table}

\section{Facial part segmentation experiment} \label{sec:face_segmentation}

\begin{figure}[!htb]
	\begin{center}
		\centering
		\includegraphics[height=1.65in,trim={0.5cm 0.5cm 0.5cm 0.5cm},clip]{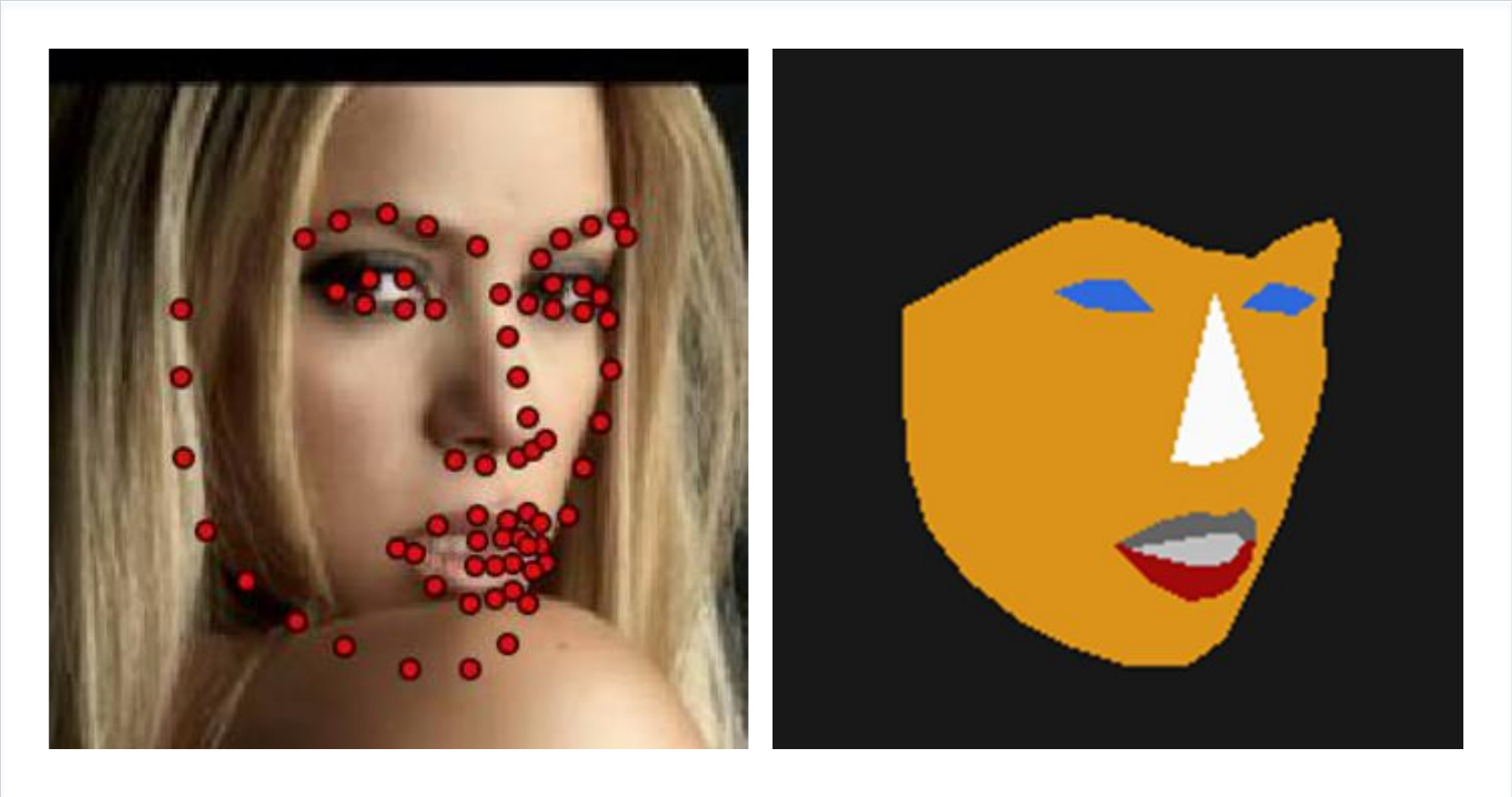}
	\end{center}
	\caption{Example of a ground truth mask (right) produced by joining the 68 ground truth keypoints (left). Each color denotes one of the seven classes.}
	\label{fig:seg_gt_example}
\end{figure}

To show that the proposed block generalizes well, producing consistent results across various datasets and tasks, in this section, we report the results of an experiment on semantic facial part segmentation. To this end, we constructed a dataset for facial part segmentation by joining together the 68 ground truth keypoints (originally provided for face alignment) to fully enclose each facial component. In total, we created seven classes: skin, lower lip, upper lip, inner mouth, eyes, nose and background. Fig.{~\ref{fig:seg_gt_example}} shows an example of a ground truth  mask.

In particular, we trained the network on the 300W dataset (approx. 3000 images) and tested it on the 300W competition testset, both Indoor\&Outdoor subsets (600 images), using the same procedure described in Section 7.

\begin{figure*}[!t]
	\centering
	\includegraphics[height=5.0in,trim={0.5cm 0.0cm 0.5cm 0.5cm},clip]{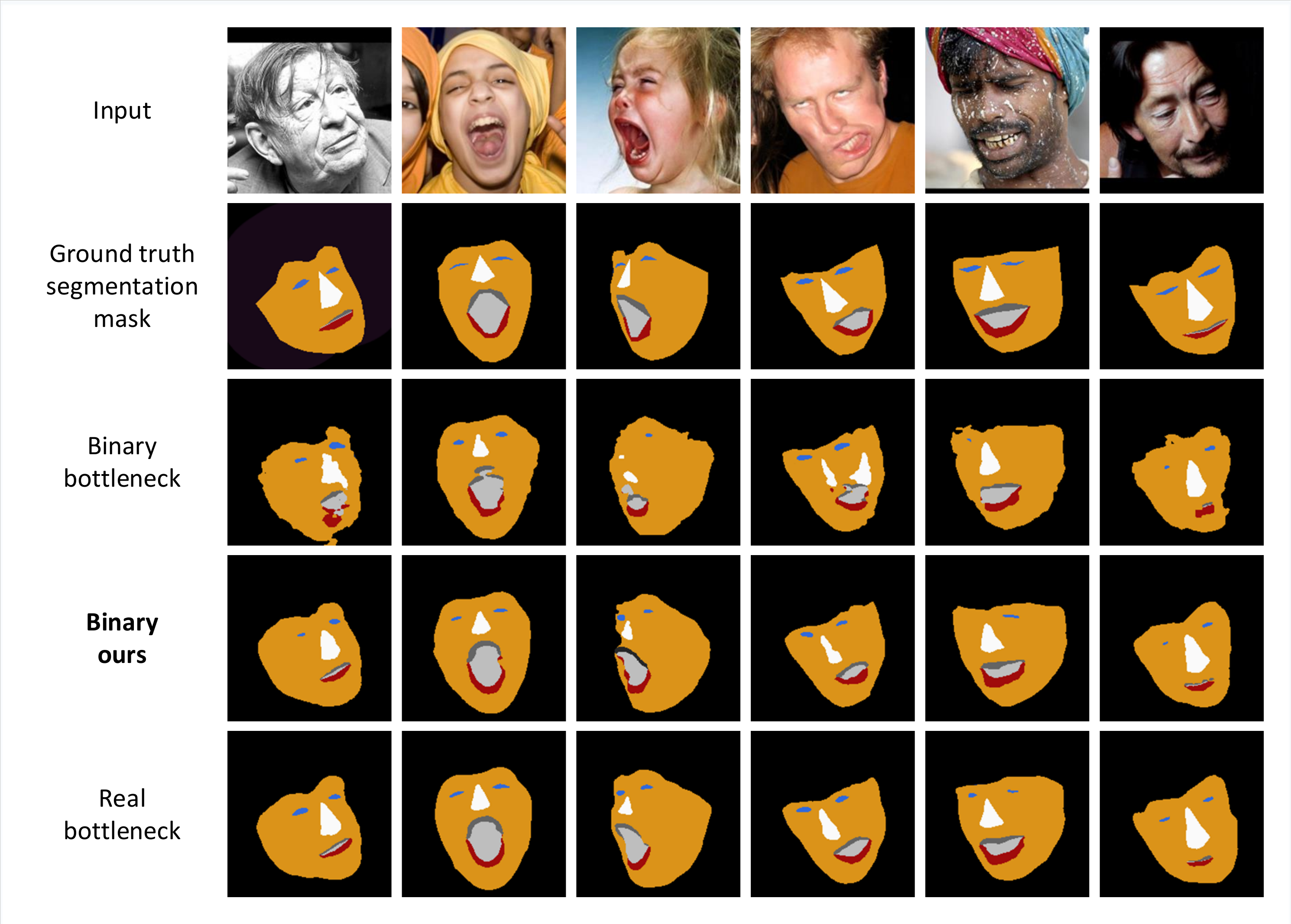}
	\caption{Qualitative results on 300W (Indoor\&Outdoor). Observe that the proposed binarized network significantly outperforms the original binary one, almost matching the performance of the real-valued network.}
	\label{fig:examples_segm}
\end{figure*}

\textbf{Architecture}. We reused the same architecture for landmark localization, changing only the last layer in order to accommodate the different number of output channels (from 68 to 7). We report results for three different networks of interest: (a) a real-valued network using the original bottleneck block (called ``Real, Bottleneck''), (b) a binary network using the original bottleneck block (called ``Binary, Bottleneck''), and (c) a binary network using the proposed block (called ``Binary, Ours''). To allow for a fair comparison, all  networks have a similar number of parameters and depth. For training the networks, we used the LogSoftmax loss {\cite{long2015fully}}. 
\newline \textbf{Results.}  Table{~\ref{tab:segm_results}} shows the obtained results. Similarly to our human pose estimation and face alignment experiments, we observe that the binarized network based on the proposed block significantly outperforms a similar-sized network constructed using the original bottleneck block, almost matching the performance of the real-valued network. Most of the performance improvement is due to the higher representation/learning capacity of our block, which is particularly evident for difficult cases like unusual poses, occlusions or challenging lighting conditions. For visual comparison, see  Fig.{~\ref{fig:examples_segm}}. 
\begin{table}[!htbp]
	\begin{center}
		\begin{tabular}{|l|c|c|c|}
			\hline
			Network type  & pixel acc. & mean acc. & mean IU\\
			\hline\hline
			Real, bottleneck   & 97.98\% & 77.23\% & 69.29\%\\
			Binary, bottleneck    & 97.41\% & 70.35\% & 62.49\%\\
			\textbf{Binary, Ours}  & 97.91\% & 76.02\% & 68.05\%\\
			\hline
		\end{tabular}
	\end{center}
	\caption{Results on 300W (Indoor\&Outdoor). The pixel acc., mean acc. and mean IU are computed as in {\cite{long2015fully}}.}
	\label{tab:segm_results}
\end{table}

\section{Visual results} \label{sec:visual}

This section provides qualitative results for our human pose estimation, face alignment and facial part segmentation experiments.

\begin{figure*}[!t]
	\centering
	\begin{subfigure}[t]{1\textwidth}
		\centering
		\includegraphics[height=2.2in,trim={0.5cm 0.5cm 0.5cm 0.5cm},clip]{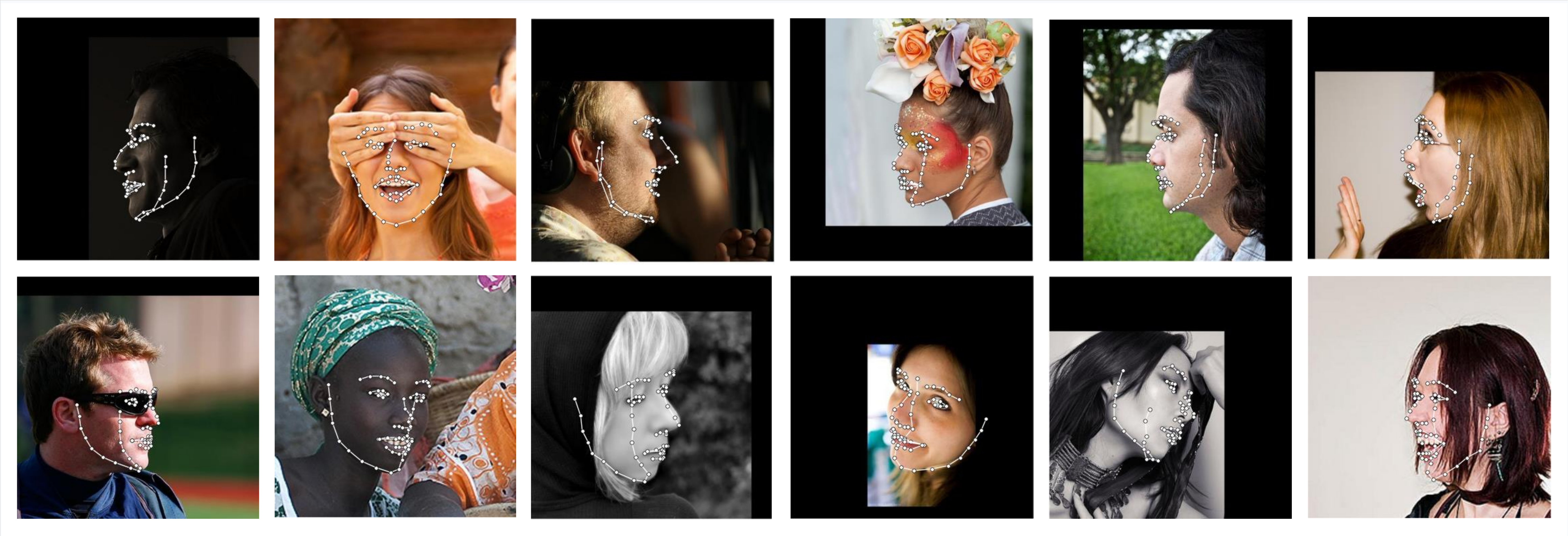}
		\caption{Fitting examples produced by our binary network on AFLW2000-3D dataset. Notice that our method copes well with extreme poses and facial expressions and lighting conditions.}
		\label{fig:examples_face}
	\end{subfigure}%
	
	\begin{subfigure}[t]{1\textwidth}
		\centering
		\includegraphics[height=2.2in,trim={0.5cm 0.5cm 0.5cm 0.5cm},clip]{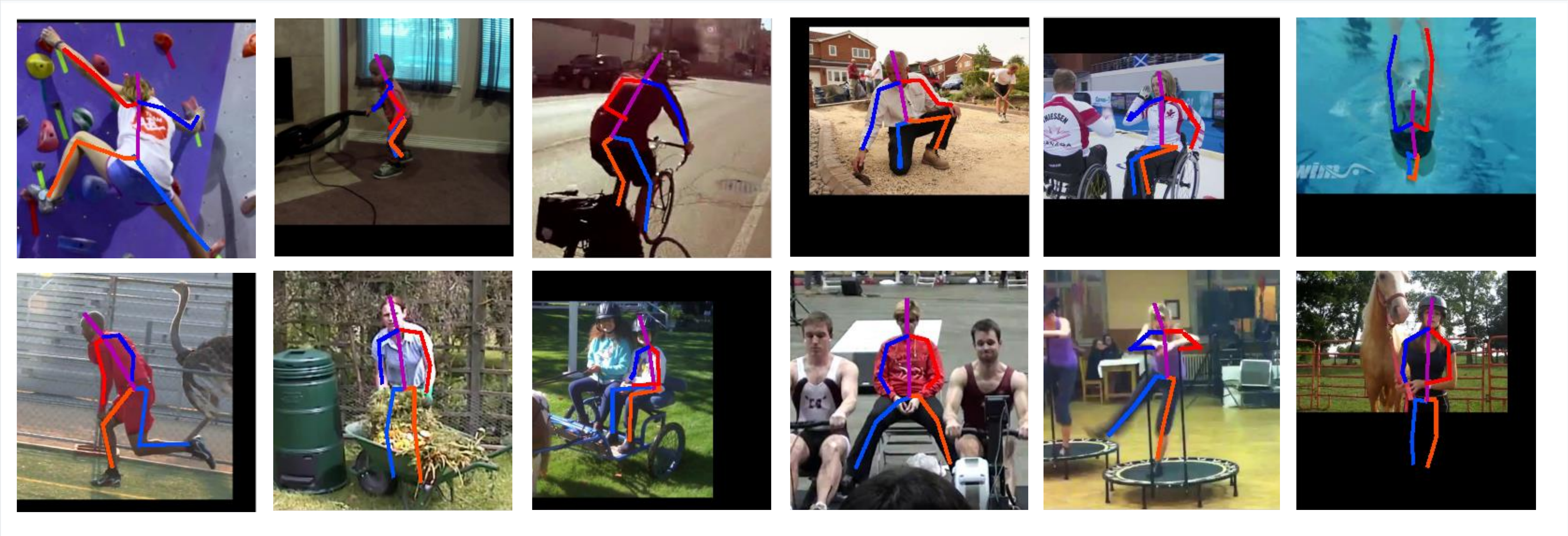}
		
		\caption{Examples of human poses obtained using our binary network. Observe that our method produces good results for a wide variety of poses and occlusions.}
		\label{fig:examples_human}
	\end{subfigure}  
	\caption{Qualitative results produced by our method on (a) AFLW2000-3D and (b) MPII datasets.}
	\label{fig:examples}
\end{figure*}

\end{document}